\documentclass[letterpaper]{article} 
\usepackage{aaai2026}  
\usepackage{times}  
\usepackage{helvet}  
\usepackage{courier}  
\usepackage[hyphens]{url}  
\usepackage{graphicx} 

\usepackage{amsmath}
\usepackage{amssymb}
\usepackage{booktabs, tabularx}
\usepackage{enumitem}
\usepackage{makecell}
\usepackage{multirow}
\usepackage[ruled,vlined]{algorithm2e}

\graphicspath{{figs/}}

\SetAlgoLined
\SetKwInput{KwInput}{Input}
\SetKwInput{KwOutput}{Output}
\SetKwComment{tcp}{$\triangleright$~}{}
\SetCommentSty{textnormal}

\usepackage{natbib}  
\usepackage{caption} 
\usepackage{newfloat}
\usepackage{float}

\usepackage{listings}
\DeclareCaptionStyle{ruled}{labelfont=normalfont,labelsep=colon,strut=off} 
\floatstyle{ruled}
\newfloat{listing}{tb}{lst}{}
\floatname{listing}{Listing}
\title{MIRAGE: Scaling Test-Time Inference\\ with Parallel Graph-Retrieval-Augmented Reasoning Chains
}
\author{
  Kaiwen Wei\equalcontrib\textsuperscript{\rm 1},
  Rui Shan\equalcontrib\textsuperscript{\rm 1},
  Dongsheng Zou\textsuperscript{\rm 1}\thanks{Corresponding author.},
  Jianzhong Yang\textsuperscript{\rm 1},\\
  Bi Zhao\textsuperscript{\rm 1},
  Junnan Zhu\textsuperscript{\rm 2},
  Jiang Zhong\textsuperscript{\rm 1}
}

\affiliations{
  \textsuperscript{\rm 1}College of Computer Science, Chongqing University, Chongqing, China\\
  \textsuperscript{\rm 2}State Key Laboratory of Multimodal Artificial Intelligence Systems, Institute of Automation, CAS, Beijing, China\\
  \{weikaiwen, dszou, zjstud\}@cqu.edu.cn,\;
  \{ruishan, yangjianzhong, zhaobi\}@stu.cqu.edu.cn,\;
  junnan.zhu@nlpr.ia.ac.cn
}

\begin{document}

\maketitle

\begin{abstract}

Large reasoning models (LRMs) have shown significant progress in test-time scaling through chain-of-thought prompting. 
Current approaches like search-o1 integrate retrieval augmented generation (RAG) into multi-step reasoning processes but rely on a single, linear reasoning chain while incorporating unstructured textual information in a flat, context-agnostic manner. As a result, these approaches can lead to error accumulation throughout the reasoning chain, which significantly limits its effectiveness in medical question-answering (QA) tasks where both accuracy and traceability are critical requirements.
To address these challenges, we propose MIRAGE (Multi-chain Inference with Retrieval-Augmented Graph Exploration), a novel test-time scalable reasoning framework that performs dynamic multi-chain inference over structured medical knowledge graphs. Specifically, MIRAGE 1) decomposes complex queries into entity-grounded sub-questions, 2) executes parallel inference chains, 3) retrieves evidence adaptively via neighbor expansion and multi-hop traversal, and 4) integrates answers using cross-chain verification to resolve contradictions. 
Experiments on three medical QA benchmarks (GenMedGPT-5k, CMCQA, and ExplainCPE) show that MIRAGE consistently outperforms GPT-4o, Tree-of-Thought variants, and other retrieval-augmented baselines in both automatic and human evaluations. Additionally, MIRAGE improves interpretability by generating explicit reasoning chains that trace each factual claim to concrete chains within the knowledge graph, making it well-suited for complex medical reasoning scenarios.
\end{abstract}

\begin{links}
    \link{Code}{https://github.com/Despacitobei/MIRAGE/}
\end{links}

\section{Introduction}

\begin{figure}[t]
\centering
\includegraphics[width=\linewidth]{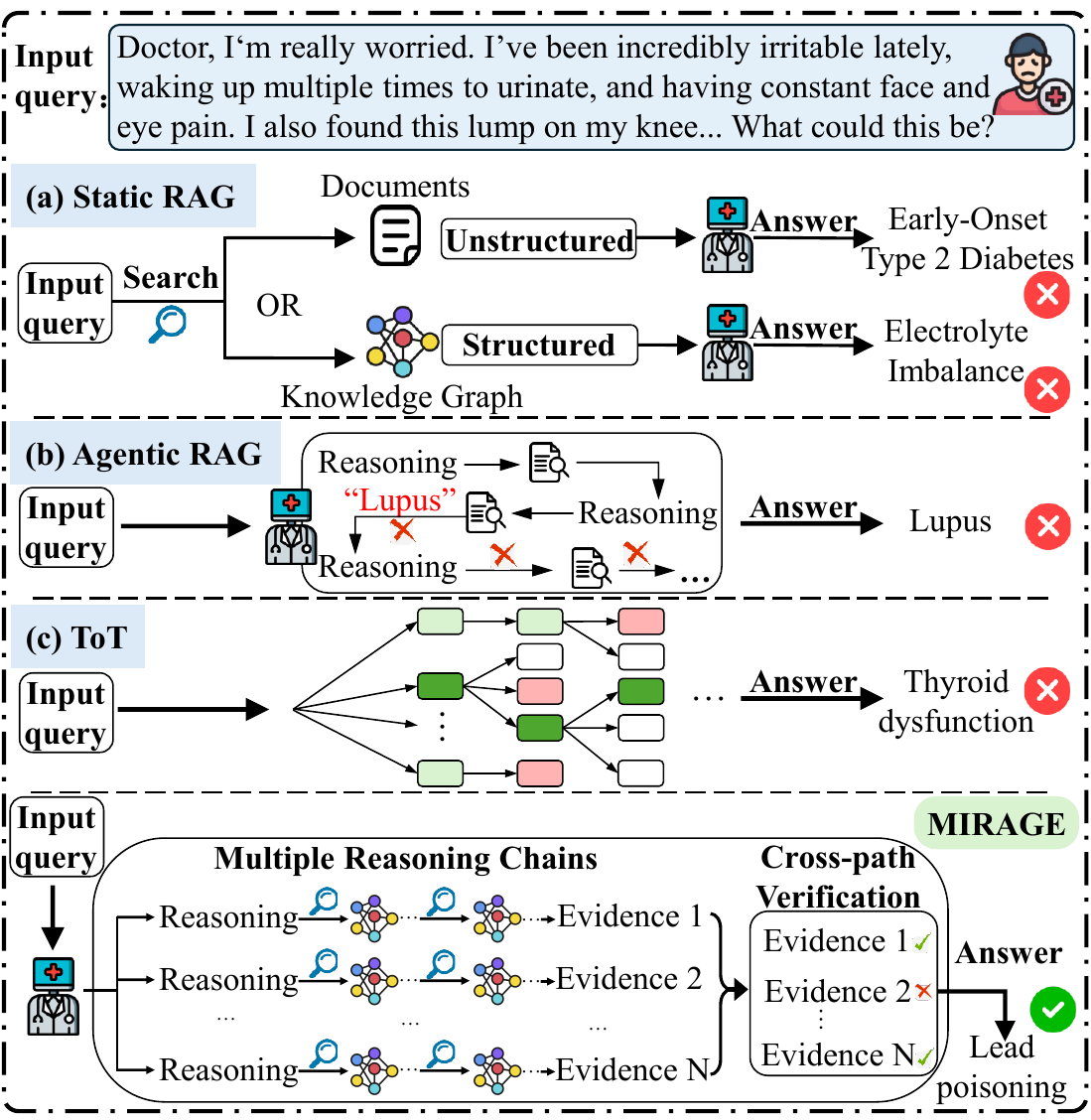}
\caption{Comparison of medical QA strategies: (a) Static RAG retrieves documents or knowledge graph entries without explicit reasoning; (b) Agentic RAG methods like Search-o1 integrate retrieval with linear reasoning; (c) ToT explores multiple reasoning chains via sampling; (d) The proposed MIRAGE combines these approaches by performing graph-based retrieval across parallel reasoning chains.}
\label{fig:comparison}
\end{figure}

The developments of large language models (LLMs) have led to the emergence of large reasoning models (LRMs), which exhibit strong multi-step reasoning abilities through chain-of-thought (CoT) prompting~\cite{10.5555/3600270.3602070}. Unlike conventional LLMs that often provide shallow or heuristic answers, LRMs adopt a slow-thinking paradigm that enables the progressive unfolding of reasoning steps. More importantly, recent advances in test-time scaling~\cite{DBLP:journals/corr/abs-2408-03314} have demonstrated that reasoning capability can be significantly enhanced at inference time without retraining the underlying model, simply by allocating more computational resources during the reasoning process. This paradigm is exemplified by models such as OpenAI's o1~\cite{openai2024openaio1card}, Qwen-QwQ~\cite{qwq-32b-preview}, and DeepSeek-R1~\cite{DBLP:journals/corr/abs-2501-12948}, which achieve substantial performance improvements through extended reasoning chains and explicit step-by-step thought processes~\cite{team2024deepseek}. Building on these foundations, agentic frameworks like Search-o1~\cite{li2025searcho1agenticsearchenhancedlarge} further enhance test-time scaling by integrating retrieval-augmented generation (RAG) into the reasoning loop, enabling models to dynamically access external knowledge sources during inference.

Despite the promise of test-time scaling, current approaches face a fundamental limitation in their scaling strategy: they primarily rely on linear expansion through sequential reasoning chains or iterative retrieval rounds. As illustrated in Figure~\ref{fig:comparison}~(b), this linear scaling approach is inherently inefficient and sensitive to error propagation. When early reasoning steps are incorrect or based on incomplete evidence, the entire extended reasoning chain becomes compromised. While techniques such as Tree-of-Thoughts (ToT)~\cite{10.5555/3666122.3666639} and its extensions like ARise~\cite{zhang2025arise} attempt to explore multiple reasoning paths, they lack coherent mechanisms for coordinating parallel inference chains and enabling explicit cross-chain verification. This limitation becomes particularly pronounced in complex domains like medicine, where reasoning errors can have critical consequences and where the linear scaling paradigm fails to effectively leverage the additional computational budget.

Furthermore, current test-time scaling methods face a critical challenge in how they expand their knowledge coverage during inference. Existing retrieval-augmented approaches~\cite{li2025searcho1agenticsearchenhancedlarge} typically acquire unstructured textual information and integrate it into the reasoning process in a flat, context-agnostic manner. This approach to knowledge scaling overlooks the inherent structural relationships and semantic hierarchies within domain knowledge, particularly in specialized fields like medicine, where understanding often depends on complex inter-entity relationships, causal chains, and hierarchical taxonomies~\cite{fang-etal-2020-hierarchical,wen-etal-2024-mindmap,wei-etal-2021-trigger}. As a result, even when more computational resources are allocated to retrieve additional information, the flat integration of isolated text fragments limits the system's ability to perform precise multi-hop reasoning and construct coherent knowledge-grounded arguments. This structural knowledge scaling challenge significantly constrains the effectiveness of test-time scaling in knowledge-intensive domains.

To address these fundamental limitations in test-time scaling, we propose Multi-chain Inference with Retrieval-Augmented Graph Exploration (MIRAGE), a novel reasoning framework that reimagines how computational resources are allocated during inference. Rather than scaling through linear chain extension, MIRAGE implements parallel scaling by decomposing complex queries into semantically coherent sub-questions and executing multiple reasoning trajectories simultaneously. This parallel inference paradigm enables more efficient utilization of computational resources while providing natural mechanisms for cross-chain verification and error correction. To address the knowledge scaling challenge, MIRAGE integrates an adaptive retrieval-reasoning loop that interacts with structured domain knowledge graphs, enabling dynamic exploration of semantic relationships through neighbor expansion and multi-hop traversal strategies. By organizing knowledge retrieval around graph structures rather than flat text fragments, MIRAGE can perform more sophisticated reasoning that leverages the inherent hierarchical and relational properties of domain knowledge. Following the evaluation protocol established by MindMap~\cite{wen-etal-2024-mindmap}, we conduct extensive experiments on three medical question answering (QA) benchmarks, demonstrating that MIRAGE consistently outperforms strong baselines including GPT-4o, Tree-of-Thought variants, and other retrieval-augmented approaches in both automatic metrics and human evaluations across all datasets.
In summary, the contributions of this paper are as follows: 

(1) We propose MIRAGE, a novel test-time scaling framework that shifts from linear chain extension to parallel multi-chain inference, enabling more efficient utilization of computational resources while providing mechanisms for cross-chain verification and error correction. 

(2) We develop a structured knowledge scaling approach through adaptive graph-based retrieval, which dynamically explores semantic relationships and hierarchical structures during inference rather than relying on flat text integration. 

(3) We conduct comprehensive evaluations on three medical QA benchmarks, demonstrating significant improvements over existing reasoning and retrieval baselines, with detailed analysis showing the effectiveness of both parallel reasoning scaling and structured knowledge scaling components in improving performance.

\begin{figure*}[t!]
\centering
\includegraphics[width=\textwidth]{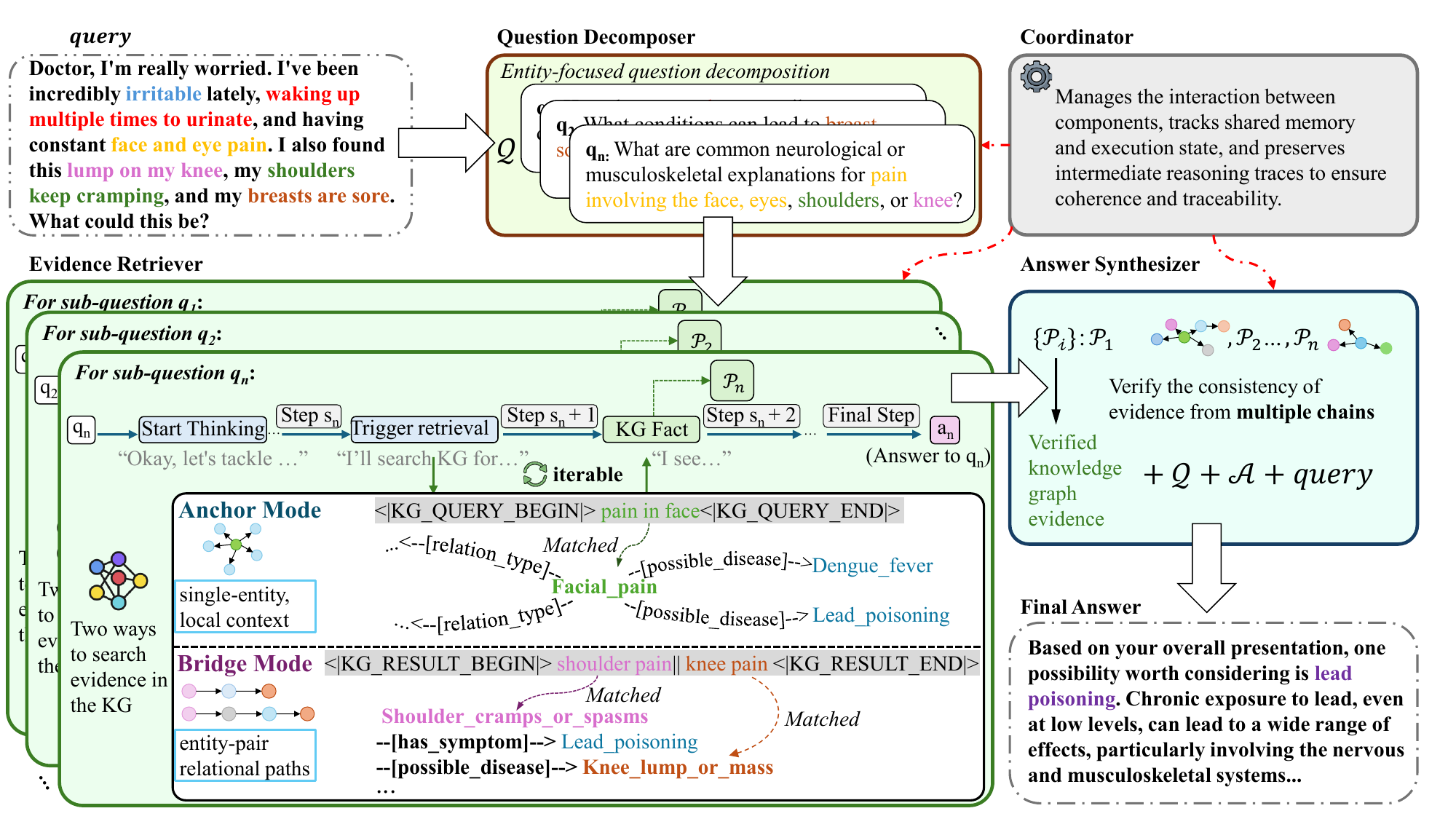}
\caption{Overview of the proposed MIRAGE framework. Given a clinical query, the system decomposes it into sub-questions, each initiating a reasoning chain. For each chain, the system iteratively retrieves knowledge graph evidence using either Anchor mode or Bridge mode. Retrieved results are coordinated and aggregated to generate the final answer.}
\label{fig:overall-framework}
\end{figure*}

\section{Related Work}

\paragraph{LLMs for Reasoning}
Prompt-based reasoning methods, including Chain-of-Thought (CoT) prompting~\cite{10.5555/3600270.3602070}, Self-Consistency~\cite{wang2023selfconsistency}, Tree-of-Thought~\cite{10.5555/3666122.3666639}, Graph-of-Thought~\cite{Besta_Blach_Kubicek_Gerstenberger_Podstawski_Gianinazzi_Gajda_Lehmann_Niewiadomski_Nyczyk_Hoefler_2024}, and Chain-of-Specificity~\cite{wei-etal-2025-chain}, have substantially improved the reasoning capabilities of large language models (LLMs) without altering model parameters~\cite{chu-etal-2024-navigate,DBLP:journals/corr/abs-2407-11511}. These methods rely on carefully designed prompts to elicit multi-step, interpretable rationales and have been extended to diverse tasks and modalities in various domains~\cite{zhang2024multimodalchainofthoughtreasoninglanguage}.

Recent advances in large reasoning models (LRMs) have led to the development of large language models that are explicitly trained to internalize multi-step reasoning processes. Representative models include DeepSeek-R1~\cite{DBLP:journals/corr/abs-2501-12948}, OpenAI-o1~\cite{openai2024openaio1card}, and Qwen-QwQ~\cite{qwq-32b-preview}. Through reinforcement learning and reasoning-oriented supervision, these models acquire explicit chain-of-thought abilities and demonstrate strong performance on complex benchmarks such as mathematics and coding~\cite{qu2025surveyefficientreasoninglarge}. Nevertheless, both prompt- and model-based approaches are limited by knowledge gaps and error accumulation, often yielding unreliable answers~\cite{chu-etal-2024-navigate,DBLP:journals/corr/abs-2407-11511,qu2025surveyefficientreasoninglarge}. Search-o1~\cite{li2025searcho1agenticsearchenhancedlarge} and its variants~\cite{DBLP:journals/corr/abs-2504-21776,guan2025deepragthinkingretrievestep} integrate agentic search into the o1-like reasoning process of LLMs but follow a monolithic, sequential paradigm lacking modularity and cross-step coordination. ARise~\cite{zhang2025arise} adds multi-path tree search with dynamic retrieval but lacks structured path alignment and global consistency, limiting complex multi-hop reasoning.

\paragraph{Retrieval-Augmented Generation} 
Retrieval-Augmented Generation (RAG) integrates external knowledge retrieval into the generation process to enhance factual accuracy and reduce hallucinations~\cite{10.5555/3495724.3496517}. Classic RAG methods typically employ static, single-step retrieval prior to generation and have been widely adopted in QA, dialogue, and summarization tasks~\cite{10.5555/3524938.3525306, zhu-etal-2025-tableeval, wei-etal-2025-chartmind}. To improve retrieval relevance, later studies explored query rewriting~\cite{ma-etal-2023-query}, reranking~\cite{glass-etal-2022-re2g}, document compression~\cite{xu2024recomp}, and  GraphRAG~\cite{DBLP:journals/corr/abs-2404-16130,wen-etal-2024-mindmap}. Meanwhile, generation-side efforts included improved conditioning strategies~\cite{10.5555/3648699.3648950} and post-generation provenance verification and evaluation~\cite{sankararaman-etal-2024-provenance,zhu-etal-2025-trove}.

Although existing RAG methods have demonstrated significant success, they remain limited in multi-step reasoning settings, where information needs to evolve dynamically and cannot be satisfied by a single retrieval step. Recent agentic RAG frameworks~\cite{li2025searcho1agenticsearchenhancedlarge,yao2023react,nakano2022webgptbrowserassistedquestionansweringhuman,asai2024selfrag,zhang2025arise} address this by interleaving retrieval and generation in a loop, enabling models to query external sources based on intermediate reasoning states. 
However, most systems still operate on unstructured text, lacking dynamic access to structured or semantically grounded representations\cite{wei-etal-2023-guide}. This limits their ability to support fine-grained, multi-hop reasoning, especially in the medical domain that requires entity-centric, relational, and hierarchical knowledge modeling.

\section{Method}
\subsection{Task Definition}
In this work, we focus on leveraging structured knowledge graphs to support multi-chain inference for medical question answering (QA). Specifically, let $query$ be a natural-language medical question posed by a clinician or patient, and let $\mathcal{G} = (\mathcal{E}, \mathcal{R})$ denote a medical knowledge graph, where $\mathcal{E}$ is a set of medical entities (e.g., diseases, drugs, symptoms, foods, ...) and $\mathcal{R}$ is a set of typed relations capturing associations between these entities. We use $\mathcal{G}$ as the primary external knowledge source because its structured format, consisting of deduplicated entities and clearly defined relations, provides concise and consistent factual information. This helps mitigate the ambiguity and redundancy often found in free-form clinical text. The system returns a single natural-language answer $a \in \mathbb{T}^{+}$, where $\mathbb{T}^{+}$ denotes the set of non-empty free-form text strings.

\label{sec:pipeline}

\subsection{Overall Framework}
\label{sec:overall-framework}

Figure~\ref{fig:overall-framework} presents the structure of MIRAGE, which comprises four main components:

\begin{itemize}[leftmargin=10pt, itemsep=2pt]
  \item \textbf{Question Decomposer}: counters the common weakness of
        earlier decomposers that emit entity-agnostic, free-form
        sub-questions; instead, it produces concise splits that are
        explicitly grounded in concrete medical entities, keeping the
        context small and noise-free for downstream steps.

  \item \textbf{Evidence Retriever}: engages in a
        think-while-search cycle, alternating language-model reasoning
        with targeted knowledge-graph queries.  Each round can refine
        or extend previous hypotheses, gradually assembling the
        evidence required for every sub-question.

  \item \textbf{Answer Synthesizer}: consolidates the partial answers
        of all sub-questions and verifies that they are mutually
        consistent, resolving any contradictions before emitting a
        single, coherent answer with traceable citations.
        
    \item \textbf{Coordinator}: manages the execution of the three components by facilitating communication through a shared in-memory workspace. It monitors the workspace and activates downstream modules as soon as their required inputs become available. 
\end{itemize}

\noindent
For example,  given the query “\textit{Why do I keep feeling fatigued even after sleeping well?}”, the question decomposer first extracts key entities like \textit{fatigue} and \textit{sleep quality} and generates sub-questions grounded in each, such as identifying causes of fatigue or conditions affecting sleep recovery. The evidence retriever then performs multi-hop reasoning over the knowledge graph, retrieving relevant entities like \textit{anemia} or \textit{thyroid disorders}. These intermediate results are merged by the answer synthesizer into a coherent explanation. Throughout the process, the coordinator monitors shared state and triggers each component as inputs become ready. Only output the final answer “\textit{Based on your symptoms, chronic fatigue syndrome is the most likely cause...}”, keeping the response concise and readable.

\subsubsection{Question Decomposition}
Question decomposition serves as the foundational step of the adaptive loop, transforming complex clinical queries into focused sub-questions that align with the granularity of \(\mathcal{G}\). This step is designed to address the limitations of prior methods, which often generate entity-agnostic sub-questions prone to noise or irrelevant retrieval. Specifically, this decomposition strategy adheres to two domain-specific principles. First, it triggers exclusively for queries involving multiple distinct medical entities (e.g., concurrent symptoms or drug-condition interactions), thereby preserving the coherence of single-topic queries and preventing unnecessary fragmentation. Second, it replaces ambiguous references with explicit entities extracted from the original query, ensuring each sub-question remains self-contained and directly mapped onto entities in \(\mathcal{E}\). This strategy is implemented via an LLM prompted with structured instructions to produce focused, entity-grounded sub-questions (see Appendix~A.1 for details).
This decomposition process initializes the loop by defining clear retrieval targets. To prevent excessive fragmentation, at most \(N_q\) sub-questions are generated per query. Each sub-question acts as a starting point for targeted knowledge graph exploration, guiding the loop to focus on relevant entities and relations rather than sprawling, unfocused searches.

\subsubsection{Graph-Augmented Evidence Retrieval}

Following decomposition, the model enters an iterative retrieval-and-reasoning loop, in which each stage is guided by structured prompts (see Appendix~A.1). The loop consists of three steps: identifying retrieval targets, querying the knowledge graph, and verbalizing the retrieved evidence. When decoding a sub-question, the model may emit a special search block \(\vartheta\), delimited by \texttt{<|KG\_QUERY\_BEGIN|>} and \texttt{<|KG\_QUERY\_END|>}, which defines the retrieval target. This block includes one or two entity mentions that are softly matched to entities in \(\mathcal{E}\) based on embedding similarity. The matched entity set $\mathcal{E}^*$ is defined as follows:
\begin{equation}
\small
\mathcal{E}^* = \{ \arg\max_{e \in \mathcal{E}} \mathrm{sim}(\hat{e}, e) \mid \hat{e} \in \hat{\mathcal{E}},\ \mathrm{sim}(\hat{e}, e) \geq \tau \}
\end{equation}

where \(\mathrm{sim}(\cdot)\) is a normalized similarity function, \(\tau\) is a threshold, and \(\hat{\mathcal{E}}\) is the set of candidate entities from \(\vartheta\). If no match exceeds the threshold, a special \texttt{no\_entity\_match} token is returned, prompting reformulation. Once the target is resolved, the model invokes the query function:
\begin{equation}
\small
\mathcal{P}_{i} = \textsc{KGSearch}(\vartheta, \mathcal{G})
\end{equation}

The strategy depends on the number of entities in the search block. For a single entity \(e\), the system retrieves its local neighborhood:
\begin{equation}
\small
\mathcal{N}(e){=}\{(e, r, e') | (e, r, e'){\in}\mathcal{G}, r{\in}\mathcal{R}\}, |\mathcal{N}(e)|{\leq}k
\end{equation}
where \(k\) is the maximum number of neighbors per relation. If two entities \((e_1, e_2)\) are specified, the system searches for typed relational chains of length up to \(h\):
\begin{equation}
\small
\mathcal{P}_h(e_1, e_2) = \{ p \mid p: e_1 \rightarrow e_2,\ |p| \leq h \}
\end{equation}
where each $p$ is a typed relational chain from $e_1$ to $e_2$.

These retrieval behaviors can be categorized into two modes: (1) \textbf{Anchor Mode} handles single-entity queries by retrieving a fixed neighborhood around the matched entity, capturing local attributes such as symptoms or treatments. It refines entity semantics while keeping sub-questions focused and clinically precise. (2) \textbf{Bridge Mode} is triggered when two entities are identified, retrieving direct or multi-hop paths connecting them. This enables cross-entity reasoning (e.g., linking symptoms to comorbidities) via intermediate biomedical relations. All queries respect relation direction and are limited to a curated set of clinically validated types. Once graph paths are retrieved, they are verbalized as:
\begin{equation}
\mathrm{Verbalize}(e_0, r, e_1) = \text{``}e_0\ r\ e_1\text{''}
\end{equation}

For example, \((\textit{Diabetes}, \textit{has\_symptom}, \textit{Fatigue})\) becomes “Diabetes has symptom Fatigue.” These fragments are inserted back into the model’s context between \texttt{<|KG\_RESULT\_BEGIN|>} and \texttt{<|KG\_RESULT\_END|>}, and logged in a shared workspace along with metadata (e.g., retrieval counts and reasoning turns). The model then continues decoding with this updated context, potentially issuing more queries if the retrieval budget allows. Over multiple rounds, the verbalized fragments accumulate into \(\mathcal{P}_i\), forming a compact, deduplicated, and directionally grounded evidence set for final answer synthesis. 

This adaptive retrieval provides three benefits: (1) it keeps context focused by injecting only relevant facts; (2) it supports iterative refinement as new evidence is retrieved; and (3) it grounds all claims in specific, traceable graph paths, enabling clinical interpretability and auditability.

\subsubsection{Answer Synthesis and Source Attribution}
Once the retrieval loop has streamed its final graph fragment, each sub-question $q_i \in \mathcal{Q}$ is associated with an answer $a_i \in \mathcal{A}$ and a corresponding chain set $\mathcal{P}_i$. At this stage, the responsibility shifts to the answer synthesizer, which revisits each $(q_i, a_i)$ pair in the context of its supporting evidence $\mathcal{P}_i$. A dedicated prompt is used to implement answer synthesis and verification (see Appendix~A.1 for details). Medical terms are normalized to their canonical synonyms, dosage units are standardized, and domain-specific rules derived from the knowledge graph are applied. For example, if a drug is found to both \emph{treat} and \emph{cause} the same symptom, the system flags the combination as biologically inconsistent.

Next, the synthesizer performs pairwise comparisons across all answers, identifying mutually exclusive diagnoses or conflicting therapeutic claims. When such conflicts arise, the system retains the answer whose supporting chain set spans a broader neighborhood of corroborating relations or aligns more closely with the original query. This majority-based verification strategy prefers answers supported by multiple independently retrieved evidence chains, while suppressing less consistent alternatives. This ensures that the final output is logically coherent and clinically sound before natural-language generation.

The reconciled set is then supplied to an LLM under a bounded prompt that forbids contradictions with verified answers but allows fallback to medical priors when the knowledge graph is silent. The model condenses the material into one or two paragraphs of patient-oriented prose explaining the likely condition, suggesting pertinent investigations, and outlining first-line management, while staying within clinically coherent bounds. Formally, using the sub-question set $\mathcal{Q}$, the corresponding answer set $\mathcal{A}$, and supporting evidence chain sets $\{\mathcal{P}_i\}$, the final reply is synthesized as:
\begin{equation}
   a \;=\; \textsc{Synth}\bigl(\textit{query},\,\mathcal{Q},\,\mathcal{A},\,\{\mathcal{P}_i\}_{q_i \in \mathcal{Q}}\bigr)
\end{equation}
where $a$ is the reply shown to the patient.

After generation, the system serializes the final response \( a \), the original \textit{query}, the validated sub-answers \((q_i, a_i)\), their supporting evidence chains \( \mathcal{P}_i \), and lightweight runtime metadata into a unified, machine-readable audit record. This record consolidates every stage of the pipeline, including decomposition, retrieval, and synthesis, and serves as a provenance-preserving trace that enhances transparency, mitigates the opacity of large language models, and supports downstream tasks such as auditing, dataset refinement, and reasoning inspection. By detecting contradictions before generation and suppressing unsupported statements, the synthesizer reduces hallucinations and reinforces clinical accuracy. The system's modular architecture also supports seamless integration of alternative language models or rule-based engines without changes to upstream retrieval components.

The overall workflow of MIRAGE is illustrated in Algorithm~1 of Appendix~B.

\section{Experiment}  

\subsubsection{Dataset}
We evaluate MIRAGE on three public medical QA benchmarks with corresponding knowledge graphs: GenMedGPT-5k with EMCKG, and CMCQA and ExplainCPE with CMCKG~\cite{wen-etal-2024-mindmap}. These datasets cover diverse clinical QA settings, including open-ended questions, multi-turn dialogues, and multiple-choice exams, across both English and Chinese. The paired knowledge graphs contain structured medical facts (e.g., diseases, symptoms, treatments) to support multi-hop reasoning. Dataset statistics are provided in Appendix~C.

\subsubsection{Evaluation}
We use both automatic and human-aligned metrics for comprehensive evaluation. For semantic similarity, we report \textit{BERTScore}~\cite{Zhang2020BERTScore}. To assess factual and clinical quality, we adopt \textit{GPT-4o Ranking}~\cite{openai2024gpt4ocard}, prompting GPT-4o to perform pairwise and listwise evaluations based on correctness, reasoning, and completeness (specific prompt in Appendix~A.2). To reduce bias, output order is randomized across trials. For pairwise comparisons, we report win/tie/loss rates. On the ExplainCPE dataset, we additionally report \textit{answer accuracy} based on ground-truth labels.

\begin{table*}[t]
\centering
\small
\setlength{\tabcolsep}{5pt}
\renewcommand{\arraystretch}{0.8}
\begin{tabular}{l ccc c ccc c ccc c c}
\toprule
\multirow{3}{*}{\textbf{Method}} 
& \multicolumn{4}{c}{\textbf{GenMedGPT-5k}} 
& \multicolumn{4}{c}{\textbf{CMCQA}} 
& \multicolumn{5}{c}{\textbf{ExplainCPE}} \\
\cmidrule(lr){2-5}
\cmidrule(lr){6-9}
\cmidrule(lr){10-14}
& \multicolumn{3}{c}{BERT-Score} & \multirow{2}{*}{Rank} 
& \multicolumn{3}{c}{BERT-Score} & \multirow{2}{*}{Rank} 
& \multicolumn{3}{c}{BERT-Score} & \multirow{2}{*}{Rank} & \multirow{2}{*}{Acc. (\%)} \\
\cmidrule(lr){2-4} \cmidrule(lr){6-8} \cmidrule(lr){10-12}
& Prec. & Rec. & F1 
& 
& Prec. & Rec. & F1 
& 
& Prec. & Rec. & F1 
& & \\
\midrule
GPT-4o             & 0.810 & 0.840 & 0.825 & 7.4 
                   & 0.839 & 0.859 & 0.849 & 7.2 
                   & 0.873 & 0.875 & 0.873 & 6.6 & 77.8\% \\
GPT-4o+ToT         & 0.825 & 0.858 & 0.841 & 5.9 
                   & 0.843 & 0.858 & 0.850 & 6.7 
                   & 0.891 & 0.872 & 0.881 & 5.8 & 80.2\% \\
QWQ-32B            & 0.816 & 0.857 & 0.836 & 4.4 
                   & 0.840 & 0.859 & 0.849 & 4.6 
                   & \underline{0.892} & 0.882 & 0.887 & 4.6 & 82.8\% \\
BM25 Retriever     & 0.805 & 0.862 & 0.832 & 4.0 
                   & 0.841 & \textbf{0.861} & 0.851 & 4.5 
                   & 0.890 & 0.878 & 0.884 & \underline{3.3} & 83.5\% \\
Embedding Retriever & 0.799 & 0.860 & 0.828 & 5.4 
                   & 0.841 & \underline{0.860} & 0.850 & 4.5 
                   & 0.890 & 0.881 & 0.885 & 5.0 & 82.4\% \\
MindMap            & 0.820 & 0.861 & 0.841 & 3.8 
                   & 0.838 & 0.858 & 0.847 & 3.1 
                   & 0.891 & \textbf{0.887} & \textbf{0.889} & \textbf{3.1} & \underline{84.6\%} \\
Search-o1          & \underline{0.832} & \textbf{0.868} & \underline{0.849} & \underline{3.3} 
                   & \textbf{0.847} & 0.857 & \underline{0.852} & \underline{3.0}
                   & 0.883 & \underline{0.886} & 0.885 & 4.6 & 80.7\% \\
\midrule
MIRAGE (ours)      & \textbf{0.841} & \underline{0.864} & \textbf{0.852} & \textbf{1.8} 
                   & \underline{0.845} & \textbf{0.861} & \textbf{0.853} & \textbf{2.8} 
                   & \textbf{0.893} & 0.884 & \underline{0.888} & \textbf{3.1} & \textbf{84.8\%} \\

\bottomrule
\end{tabular}
\caption{
Performance comparison across three medical QA datasets. 
Rank indicates the average GPT-4o Ranking (lower is better). 
Acc. (\%) is answer accuracy on ExplainCPE, which consists of multi-answer questions.
}
\label{tab:medical-results}
\end{table*}

\subsubsection{Baselines}
We compare MIRAGE with a range of baselines covering different reasoning and retrieval strategies. GPT-4o~\cite{openai2024gpt4ocard} serves as a strong general model, and GPT-4o+ToT~\cite{10.5555/3666122.3666639} augments it with Tree-of-Thought prompting for multi-step reasoning. QWQ-32B~\cite{qwq32b} represents a large reasoning model trained for end-to-end inference without prompting.
We also include retrieval-augmented QWQ-32B variants: BM25 Retriever~\cite{10.1561/1500000019} (sparse matching), Embedding Retriever~\cite{10.5555/3495724.3496517} (dense similarity), and MindMap~\cite{wen-etal-2024-mindmap}, which uses knowledge graph-based multi-hop retrieval. Search-o1~\cite{li2025searcho1agenticsearchenhancedlarge} performs dynamic document retrieval with agent-driven iterative refinement.

\subsubsection{Implementation Details}
We use the open-source Qwen-QWQ-32B~\cite{qwq32b} model as the backbone LLM for all core components of the proposed framework, including question decomposition, evidence retrieval, and answer synthesis. For the evidence retrieval and final answer generation, we set the maximum input length to 32{,}768 tokens. Please refer to Appendix~D for more details.

\subsubsection{Main Results}
Table~\ref{tab:medical-results} reports the performance of all methods across three medical QA benchmarks. We summarize the following key observations:
(1) MIRAGE consistently achieves the best overall performance in both GPT-4o ranking and answer accuracy. For instance, on GenMedGPT-5k, it reaches a rank of 1.8, and on ExplainCPE, it achieves 84.8\% accuracy, outperforming both large-scale models and retrieval-augmented baselines.
(2) QWQ-32B outperforms GPT-4o+ToT, indicating that reasoning abilities learned during pretraining may generalize better than prompt-based strategies, even with more powerful LLMs.
(3) Static retrieval methods like BM25 and dense retrievers show unstable performance, sometimes helpful but often introducing irrelevant or noisy content. In contrast, structured retrieval via knowledge graphs (e.g., MindMap) provides more consistent improvements, reflecting the need for structure-aware reasoning in medical QA.
(4) Search-o1 performs competitively on GenMedGPT-5k and CMCQA but drops notably on ExplainCPE, likely due to noisy web content. MIRAGE remains robust by relying on curated, structured medical knowledge, which enhances both accuracy and reliability.

\begin{table}[t]
\centering
\small
\setlength{\tabcolsep}{2.8pt}
\renewcommand{\arraystretch}{0.8}
\begin{tabular}{lccccc}
\toprule
\textbf{Method} & \multicolumn{5}{c}{\textbf{ExplainCPE}} \\
\cmidrule(lr){2-6}
& Prec. & Rec. & F1 & Rank & Acc.(\%) \\
\midrule
GPT-4o                      & 0.873 & 0.875 & 0.873 & 7.0 & 77.8\%  \\
GPT-4o+ToT                  & \textbf{0.891} & 0.872 & 0.881 & 4.9 & 80.2\%  \\
DeepSeek-R1-32B             & 0.889 & 0.875 & 0.882 & 4.5 & 82.2\%  \\
BM25 Retriever (DS)      & 0.888 & \underline{0.881} & \underline{0.884} & 3.5 & 83.4\%  \\
Embedding Retriever (DS) & \underline{0.890} & 0.880 & \textbf{0.885} & 4.3 & 83.3\%  \\
MindMap (DS)             & 0.887 & \underline{0.881} & \underline{0.884} & \underline{3.1} & \underline{84.1\%}  \\
Search-o1 (DS)           & 0.878 & 0.873 & 0.875 & 5.4 & 81.1\%  \\
\midrule
MIRAGE (DS)              & 0.887 & \textbf{0.883} & \textbf{0.885} & \textbf{2.9} & \textbf{84.4}\%  \\
\bottomrule
\end{tabular}

\caption{Comparison of MIRAGE and baseline methods using the DeepSeek-R1-32B backbone (denoted as DS).}
\label{tab:deepseek-results}
\end{table}

\begin{figure}[t]
    \centering
    \includegraphics[width=\linewidth]{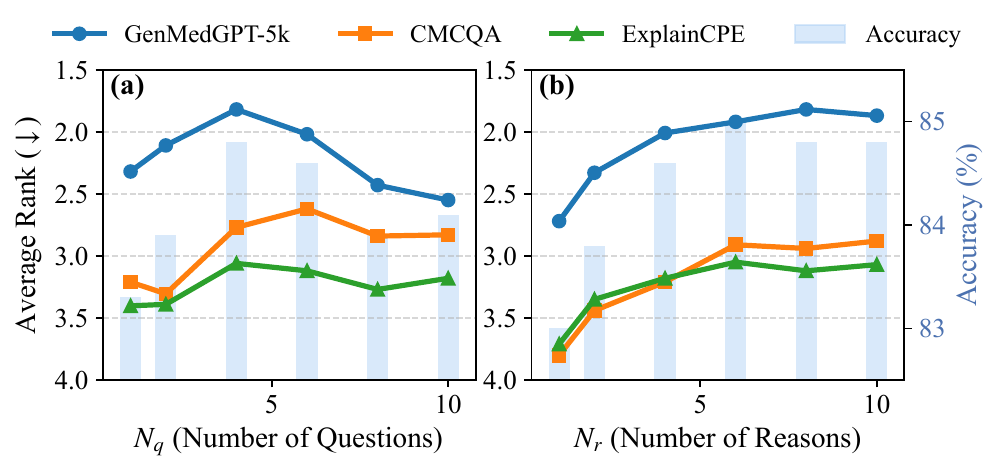}
    \caption{Effect of the decomposition threshold \(N_q\) (a) and retrieval threshold \(N_r\) (b) on GPT-4o Ranking and accuracy.}
    \label{fig:Parameter-study}
\end{figure}

\begin{figure}[t]
\centering
\includegraphics[width=0.90\linewidth]{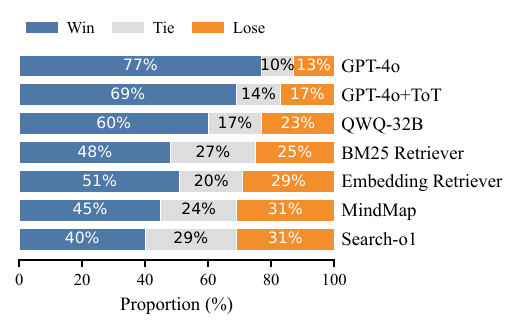}
\caption{Human evaluation results on GenMedGPT-5k.}
\label{fig:Human_evaluation}
\end{figure}

To validate the reliability of GPT-4o-based automatic evaluation, we conduct a human study on 100 randomly sampled examples from GenMedGPT-5k. We recruit 2 medical graduate students with strong English proficiency (IELTS score $\geq$ 7.0) and biomedical backgrounds, and perform pairwise comparisons between MIRAGE and all baselines, assessing factual accuracy, reasoning clarity, and clinical fluency. 
The specific metric is shown in Appendix~E. As illustrated in Figure~\ref{fig:Human_evaluation}, MIRAGE receives the highest overall preference rate, with substantial win margins and few ties or losses. These results confirm the superiority of MIRAGE from a human perspective and demonstrate strong alignment between GPT-4o rankings and human judgments.

To evaluate the generalizability of MIRAGE, we conduct an experiment on the ExplainCPE dataset using DeepSeek-R1-32B~\cite{DBLP:journals/corr/abs-2501-12948} as the backbone. All baselines use this backbone  accordingly for fair comparison. As shown in Table~\ref{tab:deepseek-results}, MIRAGE still outperforms other DeepSeek-based variants, including DeepSeek+ToT, achieving the highest GPT-4o rank and strong answer accuracy. These results confirm that MIRAGE remains effective across different backbone models, highlighting its adaptability and robustness.

\begin{figure*}[t]
  \centering
  \includegraphics[width=0.92\textwidth]{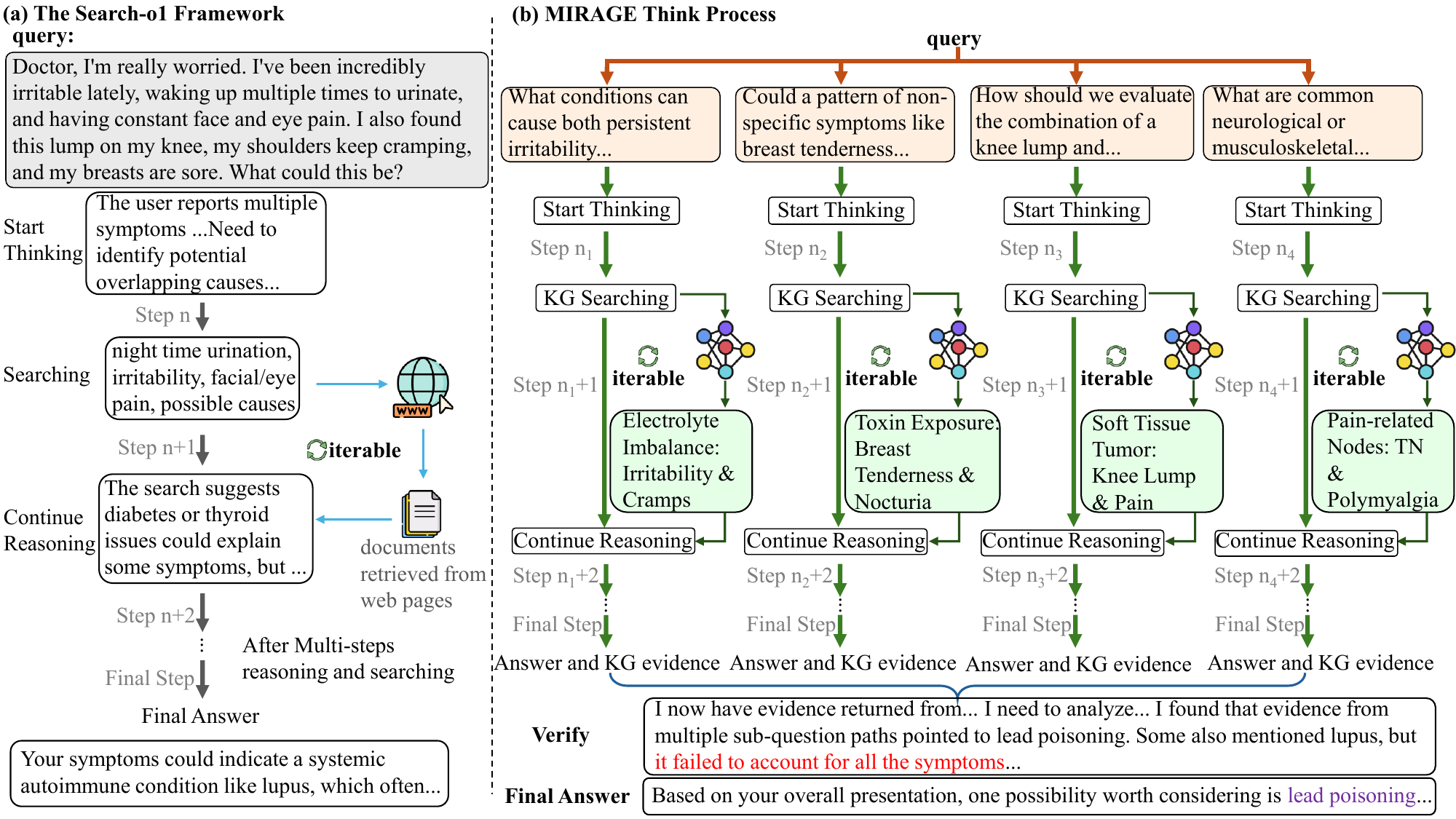}
  \caption{Case Study comparison between single-chain method and the proposed multi-chain Graph RAG reasoning method.
  }
  \label{fig:Case-study}
\end{figure*}

\begin{table*}[t]
\centering
\small
\setlength{\tabcolsep}{8pt}
\renewcommand{\arraystretch}{0.85}
\begin{tabular}{lccc ccc ccc}
\toprule
\textbf{MIRAGE vs Ablated Versions} 
& \multicolumn{3}{c}{        w/o Decomposer      } 
& \multicolumn{3}{c}{       w/o Synthesizer      } 
& \multicolumn{3}{c}{w/o Decomposer + Synthesizer} \\
\cmidrule(r){2-4} \cmidrule(r){5-7} \cmidrule(r){8-10}
\textbf{Metrics} & Win & Tie & Lose 
                & Win & Tie & Lose 
                & Win & Tie & Lose \\
\midrule
Illness identification & 49.53 & 31.13 & 19.34 & 47.64 & 36.32 & 16.04 & 52.36 & 28.77 & 18.87  \\
Treatment suggestion  & 34.43 & 54.25 & 11.32  & 41.04 & 50.47 &  8.49 & 47.17 & 47.64 &  5.19  \\
Overall correctness & 38.21 & 49.53 & 12.26  & 43.40 & 42.92 & 13.68 & 45.28 & 39.62 & 15.09  \\

\midrule
Average & 40.72 & 44.97 & 14.31 & 44.03 & 43.23  & 12.73  & 48.27  & 38.68 & 13.05  \\
\bottomrule
\end{tabular}
\caption{Pairwise comparison between ablated versions.}
\label{tab:ablation-study}
\end{table*}

\subsubsection{Ablation Study}
We perform GPT-4o-based pairwise comparisons between MIRAGE and its ablated variants, averaging over swapped output orders to mitigate positional bias. As shown in Table~\ref{tab:ablation-study}, MIRAGE consistently demonstrates a strong win--loss advantage over all ablated versions, with win rates around 40--48\% and loss rates below 15\%. While ties are common due to the conservative nature of the evaluator, the clear win margins indicate that both the Question Decomposer and Answer Synthesizer play essential roles in MIRAGE’s effectiveness.

\subsubsection{Further Analysis}
We explore the impact of the sub-question threshold \(N_q\) and retrieval threshold \(N_r\). Specifically, we vary \(N_q\) while fixing \(N_r\) at its default of 5, and vary \(N_r\) while fixing \(N_q\) at its default of 4. This analysis examines how query decomposition and evidence retrieval jointly affect multi-hop reasoning and answer quality. We report the average GPT-4o rank across all datasets and the accuracy on ExplainCPE. As shown in Figure~\ref{fig:Parameter-study}~(a), performance improves with increasing \(N_q\) up to a point, then declines as over-decomposition introduces longer or noisier reasoning chains. Although \(N_q\) only sets an upper bound, higher values often lead to overly aggressive splitting. On dialogue-style queries (e.g., CMCQA), performance may even drop from \(N_q = 1\) to \(N_q = 2\), likely due to disrupted contextual flow. In contrast, Figure~\ref{fig:Parameter-study}~(b) shows that increasing \(N_r\) yields diminishing but generally positive returns, without the sharp decline seen with \(N_q\). This suggests that the retrieval process is demand-driven: the system often performs fewer retrieval steps than the upper limit \(N_r\), stopping early when sufficient information is obtained, which helps avoid introducing unnecessary noise or redundancy. 

\subsubsection{Case Study}
To illustrate the contrast between single-chain and multi-chain graph RAG reasoning, we present a case study comparing Search-o1 and MIRAGE (Figure~\ref{fig:Case-study}). The baseline Search-o1 performs monolithic reasoning over the entire symptom set using unstructured web retrieval, often leading to information overload and vague explanations.
In contrast, MIRAGE identifies toxin exposure and neurological symptoms via separate chains and integrates them to suggest lead poisoning, effectively disentangling complex cases to produce coherent, clinically useful conclusions.

\section{Conclusion}
In this paper, we propose MIRAGE, a test-time scalable framework for dynamic multi-chain reasoning over structured medical knowledge graphs. By decomposing complex queries, adaptively retrieving structured evidence, and verifying answers across parallel inference chains, MIRAGE overcomes key limitations of existing linear and unstructured methods. 
Experiments on three medical QA benchmarks show that MIRAGE consistently improves both performance and interpretability, making it well-suited for high-stakes domains such as medicine, where accurate and traceable reasoning is essential.

\section{Acknowledgments}
This research was supported by the China National Key R\&D Program (Grant No. 2024YFC2707800, 2024YFC2707804), the National Science and Technology Major Project (2025ZD0551300, 2025ZD0551301,  2025ZD0551302), the National Natural Science Foundation of China (62176029, 62506050), the China Postdoctoral Science Foundation Funded Project (2024M763867), and the Chongqing Higher Education Teaching Reform Research Project (No. 242009).

\bibliography{aaai2026}

\end{document}